\documentclass[conference]{./support/ieeeconf}
\usepackage{times}
\usepackage{amsmath}
\usepackage[pdftex]{graphicx}
\usepackage{caption}
\usepackage{amsmath,amssymb,amsopn,amstext,amsfonts}
\usepackage{cancel}
\usepackage[space]{cite}
\usepackage{pdfsync}
\usepackage{balance}
\usepackage{color}
\usepackage{url}
\usepackage{booktabs}
\usepackage{threeparttable}
\usepackage[linkcolor=black,citecolor=black,urlcolor=black,colorlinks=true]{hyperref}
\usepackage{multirow}
\usepackage[outdir=./epstopdf/]{epstopdf}
\usepackage{graphicx}
\usepackage{array}
\usepackage{verbatim}
\usepackage{makecell}

\usepackage{threeparttable}

\usepackage{stfloats}
\usepackage{subfig}

\makeatletter  
\newif\if@restonecol  
\makeatother

\usepackage[linesnumbered,ruled,vlined]{algorithm2e}
\usepackage{algpseudocode}  
\usepackage{amsmath}  

\IEEEoverridecommandlockouts

\DeclareMathOperator*{\argminA}{arg\,min}

\graphicspath{{./figure/}}

\DeclareGraphicsExtensions{.pdf,.png,.jpg,.eps,.PNG}
\title{\LARGE \bf Cross-Modal Visual Relocalization in Prior LiDAR Maps Utilizing Intensity Textures}
\author{Qiyuan Shen$^{1}$, Hengwang Zhao$^{1}$, Weihao Yan$^{1}$, Chunxiang Wang$^{1}$, Tong Qin$^{2}$, Ming Yang$^{1,*}$
		\thanks{    This work was supported by the National Natural Science Foundation of China (U22A20100 / 62373250 / 62173228).
                        $^{1}$Department of Automation, Shanghai Jiao Tong University, Shanghai, 200240, China.
                        $^{2}$Global Institute of Future Technology, Shanghai Jiao Tong University, Shanghai, China.
		{  *Corresponding author ({e-mail: \tt\small mingyang@sjtu.edu.cn})}.
	}}

\begin{document}

\maketitle
\thispagestyle{empty}
\pagestyle{empty}

\topskip=0pt
\begin{abstract}


Cross-modal localization has drawn increasing attention in recent years, while the visual relocalization in prior LiDAR maps is less studied. Related methods usually suffer from inconsistency between the 2D texture and 3D geometry, neglecting the intensity features in the LiDAR point cloud. In this paper, we propose a cross-modal visual relocalization system in prior LiDAR maps utilizing intensity textures, which consists of three main modules: map projection, coarse retrieval, and fine relocalization. In the map projection module, we construct the database of intensity channel map images leveraging the dense characteristic of panoramic projection. The coarse retrieval module retrieves the top-K most similar map images to the query image from the database, and retains the top-K' results by covisibility clustering. The fine relocalization module applies a two-stage 2D-3D association and a covisibility inlier selection method to obtain robust correspondences for 6DoF pose estimation. The experimental results on our self-collected datasets demonstrate the effectiveness in both place recognition and pose estimation tasks.
\end{abstract}

\section{Introduction}

Visual localization is the problem of estimating the 6 Degree-of-Freedom (DoF) camera pose from which a given image was taken relative to a reference scene representation, also known as a type of map-based localization. It is a fundamental aspect of numerous applications, ranging from robotics and autonomous driving to augmented reality. 

Typically, visual localization requires consecutive camera frames or independent images to build a scene-specific 3D map using Struct-from-Motion (SfM) or Simultaneous Localization and Mapping (SLAM) methods, with the map then used to estimate the camera pose. However, pure visual methods are challenging due to the large appearance variations caused by viewpoint changes and illumination, and the visual map itself may not be accurate enough compared to natively collected 3D points. To improve visual localization performance, researchers have proposed using cross-modal information, such as LiDAR point clouds\cite{caselitz2016monocular}, making built maps more accurate and robust.

While cross-modal visual localization has drawn increasing attention\cite{10160810}\cite{chen2022i2d}, visual relocalization in prior LiDAR maps has been less studied. Compared to the task above, which can be considered local pose tracking for multi-frames, relocalization focuses on the single-frame global localization in a pre-built map. It is usually utilized in SLAM systems for global pose initialization\cite{yang2023global} and loop closure\cite{cui2022bow3d}. 

\begin{figure}[htbp]
	\centering
	\includegraphics[width=1.0\linewidth]{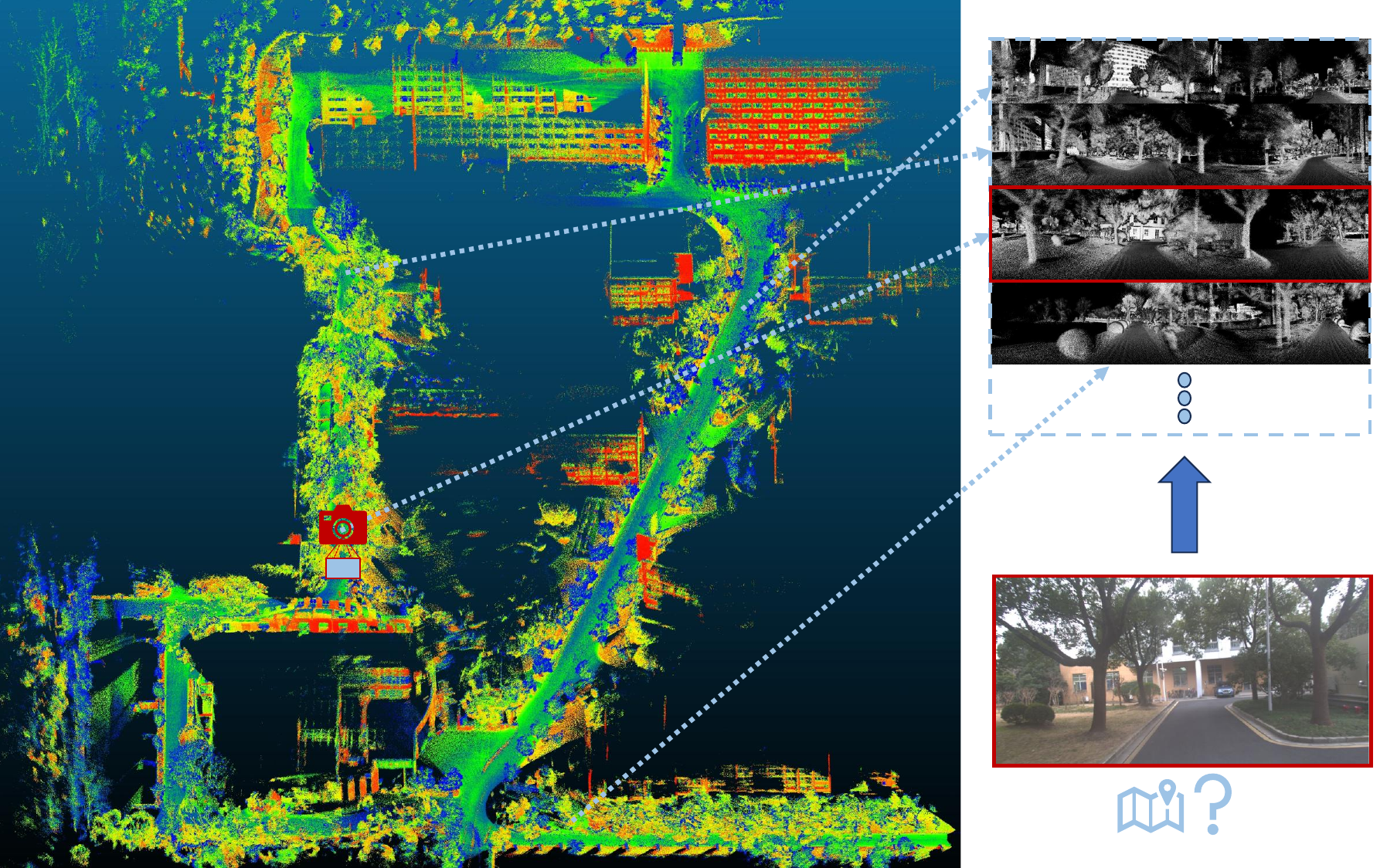}      
	\caption{The cross-modal visual relocalization in prior LiDAR maps. Given a camera image as a query, the system aims to determine its 6DoF pose in the prior LiDAR map. }
        \vspace{-0.4cm}
	\label{intro}
\end{figure}

A relocalization system hierarchically consists of two main components from coarse to fine\cite{humenberger2020robust, sarlin2019coarse}: place recognition and pose estimation. Place recognition is responsible for finding the top-K most similar place in the map to the query image. The primary obstacle lies in pose estimation, which seeks to determine the 6DoF pose of the query image relative to multiple retrieval results. The top-1 retrieval often lacks precision, leading to significant outliers. Current work in cross-modal fields mainly focuses on either image-point cloud place recognition\cite{shubodh2024lip}\cite{cattaneo2020global} or 2D-3D registration\cite{li20232d3d}, lacking a complete relocalization system contains place recognition and pose estimation together. One factor contributing to this limitation is the latter task necessitates both a pair of images and point clouds as input, rendering it unsuitable for pose estimation in relocalization scenarios.

In cross-modal association methods\cite{feng20192d3d} for place recognition and 2D-3D registration, the detect-then-match strategy involves separate feature detection in images based on texture and color and point clouds based on geometry, followed by descriptor-based matching. This method encounters challenges due to the differing natures of 2D and 3D feature detection and descriptor encoding, leading to difficulties in detecting consistent features across domains and often resulting in a low inlier ratio.

Recently, intensity has emerged as a valuable source of information in point cloud feature extraction for localization tasks. As a correlated attribute of object reflectivity, intensity serves as a texture feature within the point cloud. This attribute not only aids in LiDAR-based SLAM\cite{shan2021robust, du2023real} but also facilitates camera-LiDAR calibration\cite{koide2023general} by leveraging its consistency with image grayscale information. Given that point clouds can be projected onto a range view image, cross-modal correspondence estimation can be facilitated within a single modality by using established image registration techniques, especially with the help of powerful learning-based 2D feature extractors\cite{detone2018superpoint} and matchers\cite{lindenberger2023lightglue}.

Hence, this paper delves into the cross-modal visual relocalization, propose a hierarchical system utilizing texture consistency between the grayscale and intensity values: initially, it involves creating a map image database by leveraging the intensity channel through panoramic projection. Subsequently, the top-K similar map images are retrieved for the query camera image. Following a covisibility clustering process, two-stage 2D-3D association and covisibility inlier selection are conducted to obtain robust 2D-3D correspondences, while removing non-covisible outliers. Finally, Perspective-n-Point (PnP) and Random Sample Consensus (RANSAC) techniques are employed to estimate the 6DoF pose within the prior LiDAR map. Sufficient experiments on self-collected datasets verified the effectiveness of the proposed method with a high recall of both place recognition and pose estimation.

\section{literature review}

\subsection{Visual relocalization}

Existing visual relocalization methods typically adhere to a hierarchical localization framework \cite{humenberger2020robust}, comprising offline map construction and online localization stages. The former involves storing global and local features of keyframes and reconstructing a 3D map using Structure from Motion (SfM), while the latter conducts coarse localization via global feature matching with the database and precise localization through local feature matching, yielding 2D-3D correspondences and pose estimation.

Presently, the foremost comprehensive localization solution is HLoc \cite{sarlin2019coarse}, employing image retrieval and feature matching to achieve hierarchical localization. Moreover, another approach employs direct 2D-3D matching \cite{song2021recalling}, associating each 3D point in a pre-built 3D scene model with its corresponding image descriptor. Several methods \cite{wang2024hscnet++, do2022learning} incorporate the fusion of global and local features from the database, falling under the umbrella of learning-based direct regression. However, these approaches are highly scene-dependent and exhibit limited generalization capabilities.

Nevertheless, purely visual methods encounter numerous challenges, including variations in camera viewpoints and lighting conditions, as well as imprecisions in pre-built visual maps. A significant challenge arises from scale ambiguity in monocular vision, leading to localization results lacking real-world scale. Thus, there is a growing consideration for the adoption of LiDAR point clouds as a scene representation, supplanting visual reconstruction maps.

Currently, there are relatively few comprehensive solutions for cross-modal relocalization, with most efforts focusing solely on cross-modal place recognition. \cite{feng20192d3d} achieves hierarchical localization by simultaneously training global and local descriptors for both point clouds and images. \cite{cattaneo2020global} establishes a shared embedding space between images and LiDAR maps, facilitating image-to-point cloud place recognition. \cite{shubodh2024lip} achieves high retrieval recall accuracy by employing a popular multimodal learning approach named CLIP. However, these methods still contend with poor modality consistency, as 2D keypoint detection relies on texture information, while 3D detection is predicated on local geometry, rendering the detection of repeatable keypoints challenging.

\begin{figure*}[t]
	\centering
	\includegraphics[width=1.0\linewidth]{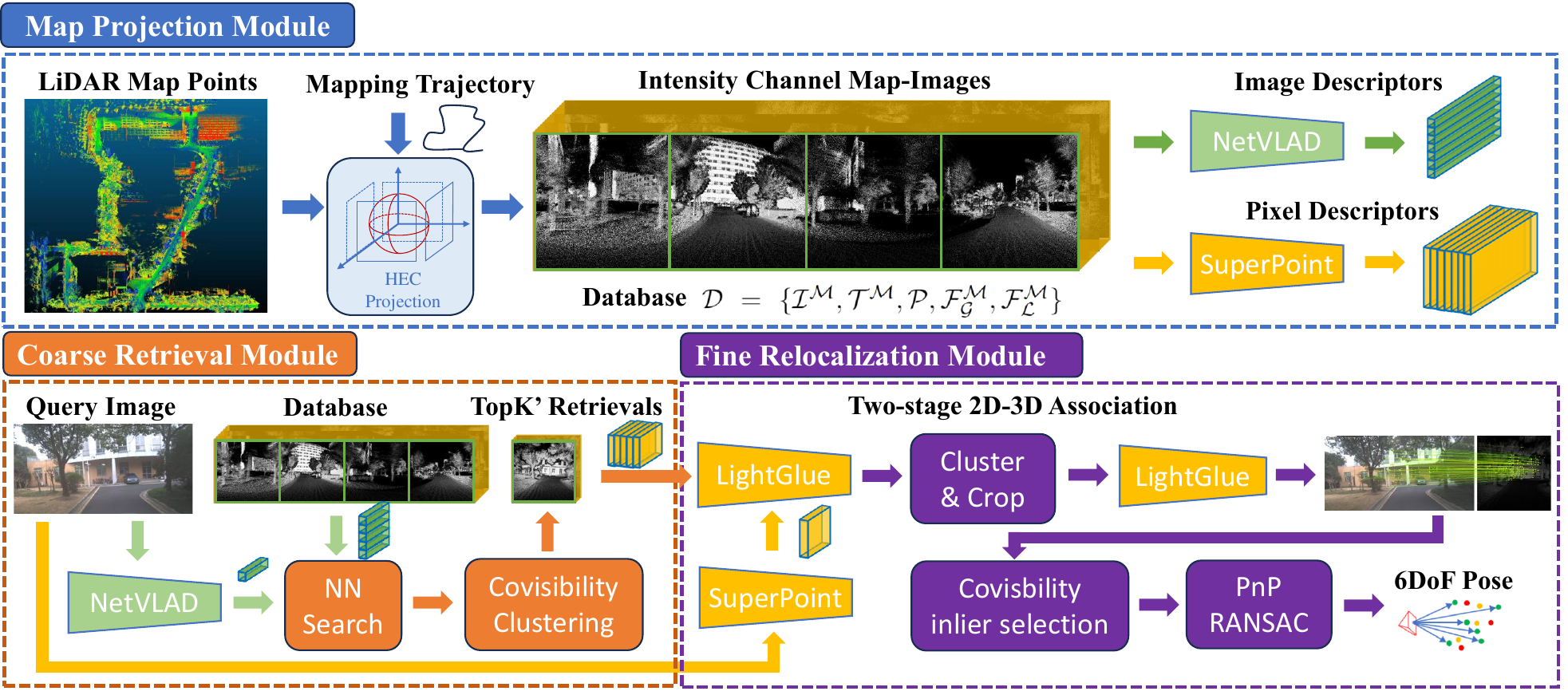}      
	\caption{The hierarchical framework of proposed cross-modal visual relocalization system in prior LiDAR maps.}
        \vspace{-0.4cm}
	\label{pipeline}
\end{figure*}

\subsection{Intensity feature extraction}

The utilization of the intensity channel within LiDAR point clouds has recently garnered significant attention as a valuable information source for point cloud feature extraction, particularly in localization tasks. Serving akin to a texture-like attribute, intensity offers supplementary insights beyond the inherent 3D geometry. This additional information proves advantageous not only in LiDAR-based SLAM but also in facilitating cross-modal data association between LiDAR and camera sensors. 

Previous works have explored LiDAR intensity for place recognition and SLAM. \cite{cop2018delight} developed a global descriptor using LiDAR intensity histograms, while \cite{wang2021intensity} offered a full SLAM system with intensity-based processing. \cite{shan2021robust} and \cite{du2023real} utilized ORB features from intensity images for loop closure and addressed geometry degeneracy in SLAM, respectively. Despite the advancements, matching intensity images poses challenges due to the commonly used spherical projection model, which correlates with the spinning LiDAR's line count. This vertical distribution is unsuitable for aligning intensity images with camera images.

In their work, \cite{cop2018delight} introduces a global feature descriptor for LiDAR intensities, comprising a set of histograms, which is subsequently employed for place recognition. Similarly, \cite{wang2021intensity} presents a comprehensive LiDAR SLAM framework incorporating both intensity-based front-end and intensity-based back-end strategies. In contrast, \cite{shan2021robust} proposes a robust LiDAR loop closure method leveraging ORB features extracted from intensity images and DBoW. Likewise, \cite{du2023real} tackles the challenge of geometry degeneracy in unstructured environments by tracking ORB features within the intensity image during front-end odometry, while simultaneously optimizing geometry residuals in the back-end. Despite the advancements, matching intensity images poses challenges due to the commonly used spherical projection model, which correlates with the spinning LiDAR's line count. This vertical distribution is unsuitable for aligning intensity images with conventional camera images.

Additionally, \cite{koide2023general} introduces a comprehensive camera-LiDAR calibration approach utilizing advanced learning-based 2D feature extractors and matchers to establish cross-modal correspondences. Their method leverages mutual information between intensity images and grayscale images to optimize extrinsic transformations. However, these methodologies do not fully exploit the consistency between intensity and grayscale values. The primary hurdle in intensity-based feature extraction is the sparsity of point cloud data, resulting in missing details at the pixel level.

\section{methodology}

\subsection{Overview}

Under the hierarchical localization framework, our cross-modal relocalization system comprises three primary modules: map projection, coarse retrieval, and fine relocalization, as depicted in Fig. \ref{pipeline}. The map projection module converts the 3D point cloud map into a panoramic image using the intensity channel, which serves as the database. Subsequently, the global and local features of each map image are stored in the database. In the coarse retrieval module, the system retrieves the top-K most similar map images to the query image from the database. To mitigate outliers among the top-K candidates, we propose a covisibility clustering method. Subsequently, the fine relocalization module estimates the 6DoF pose of the query image by establishing 2D-3D correspondences in two stages, incorporating covisibility inlier selection, and then solving the PnP problem using RANSAC. To demonstrate the generalizability of our method, we employ pre-trained state-of-the-art models for both global and local feature extraction and matching.

\subsection{Map Projection}

Imaging the point cloud plays a vital role in cross-modal registration, as it transforms the LiDAR modal into the visual-like modal for greater consistency. To achieve better texture consistency, we utilize the intensity channel of the LiDAR point cloud, a correlated attribute of reflectivity.

With the previous LiDAR map $\mathcal{M}$, the first step is to project the 3D point cloud map into a group of 2D images. To fully leverage the omnidirectional advantage of the 3D map over the 2D image, we choose to project the local map into panoramic images.
We sampled the mapping trajectories into fixed intervals to obtain projecting poses $\mathcal{T^M} = \{T^M_1, T^M_2, ..., T^M_N\}$, $\mathcal{T}\in\mathbb{R}^{N\times4\times4}$. For each projecting pose, we transformed the map points from the global coordinate $\mathcal{P}\in\mathbb{R}^{N\times M\times3}$ into the local coordinate, then projected the local maps into panoramic images $\mathcal{I^M} = \{I^M_1, I^M_2, ..., I^M_N\} $, where $\pi$ represents the projection function.  

\begin{equation}
    \mathcal{I^M} = \pi(\mathcal{T^M} \mathcal{P})
\end{equation}

\begin{figure}[t]
	\centering
        \vspace{-0.2cm}
	\captionsetup[subfloat]{}
	\subfloat[Perspective projection]{\includegraphics[height=0.32\linewidth]{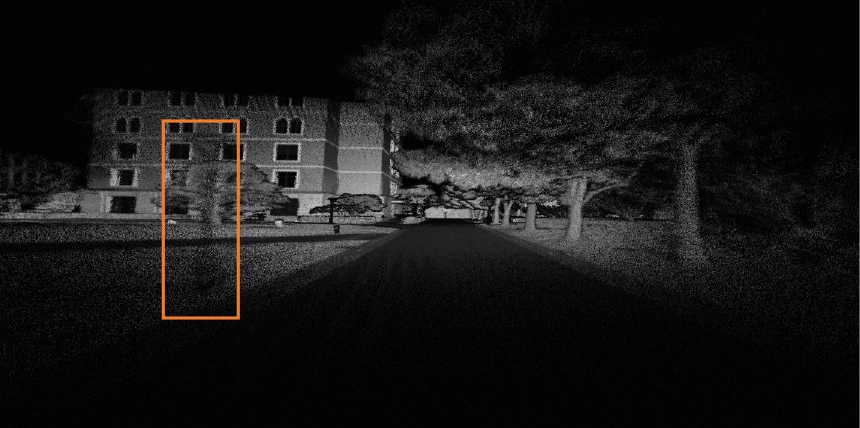}} \hfill
	\subfloat[HEC Projection]{\includegraphics[height=0.32\linewidth]{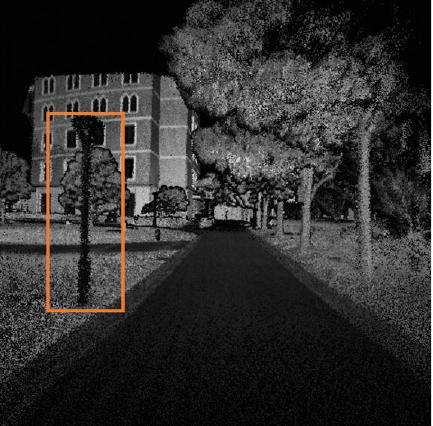}} \\
	\vspace{-0.30cm}
	\subfloat[Grayscale image]{\includegraphics[height=0.32\linewidth]{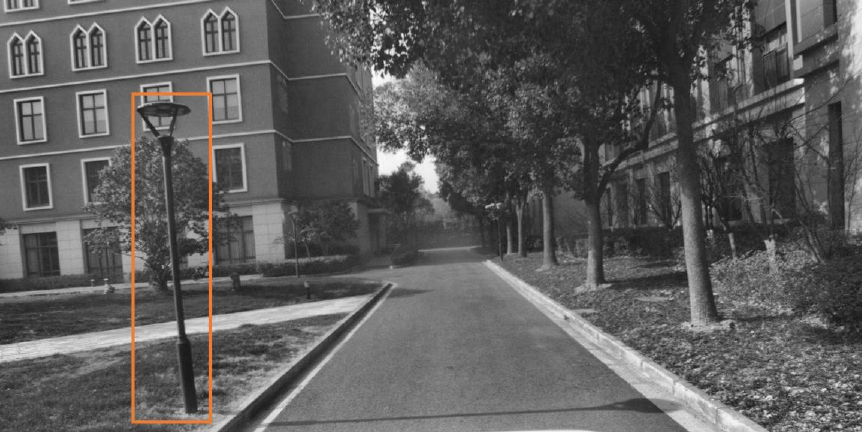}} \hfill
        \subfloat[Point Cloud]{\includegraphics[height=0.32\linewidth]{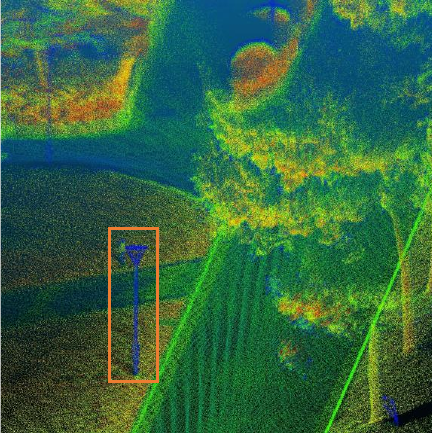}}
	\caption{The consistency comparison of different projection models between equalized map image and grayscale image. The outlined area in (a) recognized as a road lamp, is sparse while the area in (b) is much more dense which is beneficial for the following modules.}
        \vspace{-0.2cm}
	\label{fig:projection model}
\end{figure}

Recent work often utilizes an equirectangular projection model (ERP) for panoramic images, which is simple but results in relatively high distortion\cite{pi2023comprehensive}. Aside from the uneven distribution defects of the projection format, another challenge lies in missing details at the pixel level due to the sparsity of the point cloud. As depicted in Fig. \ref{fig:projection model}, the spacing between pixels projected onto the (pseudo) camera plane from closer spatial points is larger, leading to a sparse imaging effect.

To ameliorate the uniformity and the sparsity of the projection image, we use an improved method for Cube Map Projection, namely Habrid Equiangular Cube Projection (HEC) \cite{lin2019efficient}, adjusting the position of the sampled pixels on the cube corresponding to the spherical pixels, and discard the top surface and the ground surface for the sake of information validity. Given the CubeMap coordinates $(u,v)$, the HEC coordinates $(u',v')$ are calculated by Equation (\ref{tab:hec1}) and (\ref{tab:hec2}), where $t = 0.4v(u^2-1) $ is the adjusting parameter.

\begin{equation}
    u' = \frac{4}{\pi} \times \arctan u
    \label{tab:hec1}
\end{equation}

\begin{equation}
    v' = \left\{
        \begin{array}{lcl}
            v, & t=0\\
            \frac{1-\sqrt{1-4t(v-t)}}{2t}, & t\neq 0
        \end{array}
        \right.
    \label{tab:hec2}
\end{equation}

To enhance uniformity in projected map images and grayscale images, we apply a two-stage process: 3D map histogram equalization for initial normalization and CLAHE for image contrast enhancement, benefiting feature extraction and matching. For the following coarse-to-fine relocalization, the complete database $\mathcal{D} = \{\mathcal{I^M}, \mathcal{T^M}, \mathcal{P}, \mathcal{F^M_G}, \mathcal{F^M_L} \}$ stores the map images, relevant projecting poses and global map points along with their local features and the global features which will be discussed in the following sections. To be noted, the map projection module is offline preparation.

\subsection{Coarse Retrieval}

\subsubsection{Image retrieval}

Visual place recognition has been fully developed in the past decades, we choose the most classic learning-based methods NetVLAD\cite{arandjelovic2016netvlad} to extract the global feature of the query image and the map image, followed by similarity calculation. 

For query image $\mathcal{I^Q}$, we initially apply grayscale CLAHE for consistency with intensity images, then extract its global feature $\mathcal{F^Q_G}$. We calculate similarity scores against map image $\mathcal{I^M}$'s global feature $\mathcal{F^M_G}$, retrieving the top-K most similar map images. Using a sliding window strategy, we maximize similarity scores for four cube patches from the panoramic map image, aligning with the resized query image shape corresponding to the CubeMap's projected side length.

\subsubsection{Covisibility clustering}

More top-K results increase the chance of accurate matches but also the presence of outliers, impacting fine relocalization. To address this, we use covisibility clustering to refine the top-K candidates. Assuming map images with similar projection poses are covisible, we employ DBSCAN clustering based on these poses, weighted by similarity scores. Candidates are reordered by cluster size and similarity score within each cluster, selecting the top-K' from the largest cluster for further relocalization.

\subsection{Fine Relocalization}

\subsubsection{Two-stage 2D-3D association}

In the fine relocalization module, the aim is to estimate the 6DoF pose of the query image given top-K' candidates. A crucial step involves establishing 2D-3D correspondences between the query image and the map image. However, a single local feature match may not yield sufficient high-quality correspondences due to the inherent sparsity of map images and texture inconsistency between grayscale and intensity images.

The two-stage local feature matching strategy is designed to increase the number of high-quality 2D-3D correspondences by extracting the most consistent region of the query image and map image. For pairs between the query image and each map image candidate, we first conduct local feature matching and cluster the matches to obtain bounding boxes in both the query image and the map image. If the largest cluster does not contain enough matches, the pair of the query image and map image is discarded. 

\begin{figure}[t]
	\centering
	\includegraphics[width=1.0\linewidth]{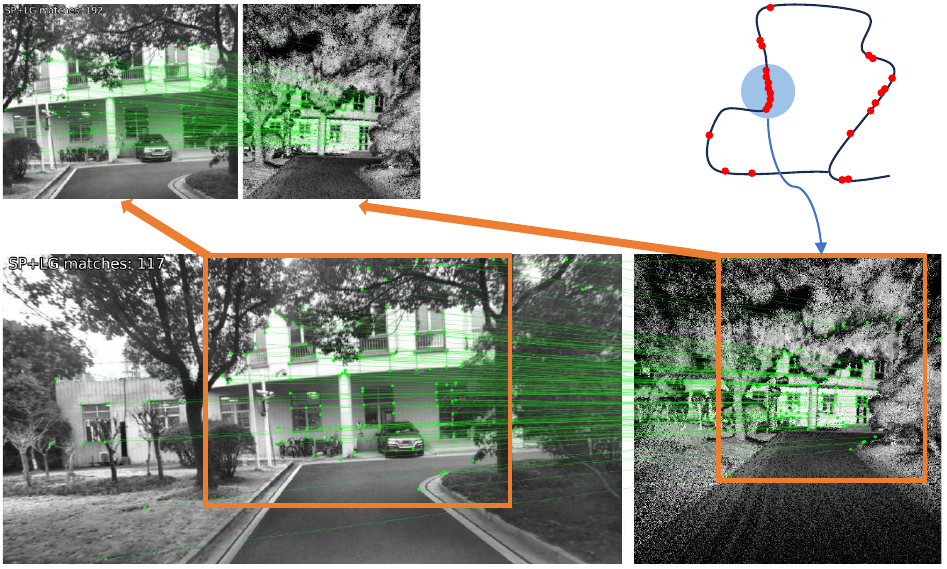}      
	\caption{The two-stage 2D-3D association results. After the covisibility clustering shown in the upper right corner, top-K' candidates are then matched with query image. The below part represents the first stage of association, while the upper left corner represents second stage results in bounding box.}
        \vspace{-0.4cm}
	\label{matching}
\end{figure}

Then we crop the bounding boxes from the original images and extract the local features again. Second-stage matching is conducted between the cropped local features, which is expected to provide more high-quality matches. The first-stage correspondences are concatenated with the second-stage correspondences to form the final 2D-2D correspondences as shown in Fig. \ref{matching}. With the correlation between the map image $\mathcal{I^M}$ and map-points $\mathcal{P}$ in the pre-built database $\mathcal{D}$, the 2D-3D correspondences are expected to be easily established given the 2D-2D correspondences. 

During the first stage, we use SuperPoint\cite{detone2018superpoint} to extract the local features of the query image $\mathcal{F^Q_L}$ and use LightGlue\cite{lindenberger2023lightglue} to match with the local feature of the retrieved map image $\mathcal{F^M_L}$. For the second stage, we use the same combination to extract more high-quality correspondences between the cropped query image and map image.

\subsubsection{Covisibility inlier selection}

In the 2D-3D association process, we implement covisibility inlier selection to further enhance correspondence quality, removing outliers with high texture similarity but from different scenes. The covisibility of the local feature points is defined as the number of map images in the database that contain the local feature points. It is assumed that the local feature points with higher covisibility are more likely to be in the same scene as the query image. We calculate the covisibility of the local feature points in the map image and filter the correspondences based on the covisibility. The correspondences with higher covisibility are retained, while the correspondences with lower covisibility are discarded.

\subsubsection{PnP with RANSAC}

Given the 2D-3D correspondences, which respectively denote the 2D keypoint coordinates $\mathcal{K^Q} = [x^Q_1, x^Q_2, ..., x^Q_J]$ and 3D coordinates $\mathcal{P} = [p^M_1, p^M_2, ..., p^M_J]$, we apply PnP with RANSAC to determine the query image's pose $\mathcal{T^Q_M}$ relative to the prior LiDAR map.
\begin{equation}
	\mathcal{T^Q_M}^* = \argminA_{\mathcal{T^Q_M}} \sum_{j=1}^{J} \rho (||\pi(\mathcal{T^Q_M}p^M_j) - x^Q_j||^2)
\end{equation}

\section{Experiments}
\subsection{Dataset}

We evaluate the proposed cross-modal visual relocalization system on our self-collected campus datasets. The dataset contains two sequences, with one scene in the college buildings and the other in the dormitory buildings. Each of these consists of two parts: a prior LiDAR point cloud map and query camera images. The point cloud data was collected using a Livox Avia, then constructed into a prior 3D map leveraging the LiDAR SLAM method. To be noted, query camera images with $1920\times960$ resolution were captured in different weather conditions a week later using a standard-fov monocular camera. To evaluate the performance of the proposed method, our dataset was annotated with ground truth camera poses, which were obtained by map-based LiDAR localization within the same prior maps.

During offline map projection, the sample intervals are set to 1m, and local maps are filtered by a max distance of 50m, while the resolution of the projection image is set to $1920\times480$ representing four $480\times480$ cubes.


\subsection{Quantitative Evaluation}

Since our system works hierarchically, we naturally evaluated in coarse retrieval procedure and fine relocalization procedure respectively. The quantitative evaluation for coarse retrieval was conducted following the standard place recognition whose evaluation metric is Recall@K: the ratio of the number of pairs that have at least one correct retrieval in the top-K retrieval results to the total number of pairs. Since our datasets are not large-scale, we consider the retrieval result with translation error below 5m instead of 20m as the correct retrieval result. Specifically, we choose the top-1, top-5, top-10. top-30 and top-50 retrieval results to evaluate the performance of the proposed method.

\begin{table}[h]
    \centering
	\setlength\tabcolsep{4pt}
    \caption{The Coarse Retrieval Results of Different Methods on SJTU Datasets.(R@K represents Recall@K)}
    \begin{tabular}{ccccccc}
    \toprule 
	 Datasets & Methods & R@1 & R@5 & R@10 & R@30 & R@50 \\ \hline
     \multirow{6}{*}{\begin{tabular}{c} College \\ 314m \end{tabular}} 
	 & LIP-Loc & 0.20 & 0.28 & 0.48 & 0.95 & \textbf{1.0}\\
	 & NetVLAD & 0.33 & 0.68 & 0.84 & 0.97 & \textbf{1.0}\\
     & Patch-NetVLAD & 0.65 & 0.91 & 0.97 & \textbf{0.99} & 0.99\\
     & Cosplace & 0.30 & 0.60 & 0.69 & 0.89 & 0.95\\
     & Eigenplaces & 0.67 & 0.81 & 0.91 & 0.94 & 0.98\\
     & Ours & \textbf{0.73} & \textbf{0.94} & \textbf{0.99} & \textbf{0.99} & \textbf{1.0}\\  \midrule
     \multirow{6}{*}{\begin{tabular}{c} Dormitory \\ 689m \end{tabular}} 
	& LIP-Loc & 0.22 & 0.34 & 0.55 & 0.80 & 0.90\\
         & NetVLAD & 0.29 & 0.60 & 0.68 & 0.87 & \textbf{0.94}\\
         & Patch-NetVLAD & 0.37 & \textbf{0.68} & 0.75 & 0.90 & 0.93\\
         & Cosplace & 0.13 & 0.39 & 0.54 & 0.66 & 0.77\\
         & Eigenplaces & 0.26 & 0.55 & 0.68 & 0.78 & 0.84\\
         & Ours & \textbf{0.39} & 0.65 & \textbf{0.82} & \textbf{0.93} & \textbf{0.94}\\ \bottomrule
    \end{tabular}

    \label{coarse}
\end{table}

The existing place recognition methods for comparison are cross-modal based LIP-loc\cite{shubodh2024lip} and visual-based methods including NetVLAD\cite{arandjelovic2016netvlad}, Patch-NetVLAD\cite{hausler2021patch}, Cosplace\cite{berton2022rethinking}, Eigenplaces\cite{berton2023eigenplaces}. Aside from LIP-Loc which uses depth channel map images, the others use intensity channel map images. Considering fairness, for comparison methods, we use their pre-trained models. For each experiment, we resize the input images to 480 to get better performance.


From the results shown in TABLE \ref{coarse}, we notice that LIP-Loc with the larger model pertained to KITTI shows bad generalization on our datasets. Our method outperforms the others including the recent state-of-the-art learning-based method Patch-NetVLAD\cite{hausler2021patch}, which can be seen as a reranking of NetVLAD results.

As for the fine relocalization evaluation, we present a metric named Relocalization Recall (RR): the ratio of queries whose translation error and rotation error between the point clouds transformed by the ground truth and the predicted transformation are below certain thresholds, in our case, 0.5m and 1\textdegree, 1m and 3\textdegree, 3m and 5\textdegree. To some extent, Relocalization Recall is capable of representing the success rate in global pose initialization and loop closure. In addition, common localization metrics like RMSE, MSE, MAE, and Max Error are also used in the quantitative evaluation.

To show the superiority of intensity textures, we compare our method with a similar hierarchical pipeline using NetVLAD for retrieval and SuperPoint with LightGlue for pose estimation while the map images are depth channels. The top-50 retrieval candidates are extracted in the coarse-retrieval module for better performance. The TABLE \ref{fine} proves the effectiveness of the proposed relocalization system. Our method shows better performance with relocalization recall and RMSE on both SJTU datasets, aligning well with our expectations. The depth images, while capturing the edge details of the foreground and background, lack the texture consistency seen in grayscale images.


\subsection{Ablation Study}

In our cross-modal relocalization system, the proposed methods share the same ultimate purpose referring to making the grayscale query image and intensity channel map image more texture-consistent, which fully utilize the potential of state-of-the-art visual pretrained models. We focus more on the applications of the proposed relocalization system and, therefore mainly show Relocalization Recall in the ablation study, as it represents the success rate to a certain extent. The results are listed in TABLE \ref{ablation}, proving the effectiveness and contributions to the whole system of proposed methods.

\begin{table}[h]
    \centering
	\setlength\tabcolsep{4pt}
    \caption{The Ablation Study On SJTU-College Datasets.}
    \begin{tabular}{ccccc}
    \toprule 
     \multirow{2}{*}{\begin{tabular}{c} Methods \end{tabular}}
	 & RR $\uparrow$ & RR $\uparrow$ & RR $\uparrow$\\
	 & (0.5m/1\textdegree) & (1m/3\textdegree) & (3m/5\textdegree) \\ \hline
	 Ours & \textbf{0.25} & \textbf{0.97} & \textbf{0.98}\\
     Ours w/o HEC projection & 0.08 & 0.61 & 0.61\\
	 Ours w/o Two-stage equalization & 0.18 & 0.92 & 0.96\\
	 Ours w/o Covisibility clustering & 0.18 & 0.92 & 0.96\\
	 Ours w/o Two-stage 2D-3D association & 0.06 & 0.92 & 0.94\\
	 Ours w/o Covisibility inlier selection & 0.24 & 0.90 & 0.94\\ \bottomrule
    \end{tabular}
    \label{ablation}
\end{table}

The HEC projection aims to enhance the density of the projected map image and two-stage equalization is applied to enhance the visualization effects of the projected map image. Covisibility clustering leverages the adjacency of projecting poses to rerank retrieval results, while two-stage 2D-3D association provides more correct 2D-3D correspondences. Before pose estimation, covisibility inlier selection rejects the local feature points that have lower observed times. 

\begin{table*}[htbp]
    \centering
	\setlength\tabcolsep{8pt}
    \caption{The Fine Relocalization Results of Different Methods On SJTU Datasets.}
    \begin{threeparttable}
    \begin{tabular}{ccccccccc}
    \toprule 
	 \multirow{2}{*}{\begin{tabular}{c} Datasets \end{tabular}} & 
     \multirow{2}{*}{\begin{tabular}{c} Methods\tnote{1} \end{tabular}} 
	 & RR $\uparrow$ & RR $\uparrow$ & RR $\uparrow$ & RMSE\tnote{2} $\downarrow$ & MSE\tnote{2} $\downarrow$ & MAE\tnote{2} $\downarrow$ & Max Error\tnote{2} $\downarrow$\\
	 && (0.5m/1\textdegree) & (1m/3\textdegree) & (3m/5\textdegree) & (m) & (m) & (m) & (m)\\ \hline
     \multirow{2}{*}{\begin{tabular}{c} College \end{tabular}} 
	 & SP+LG(Depth) & 0 & 0.11 & 0.24 & 1.41 & 1.98 & 0.98 & \textbf{4.17}\\
     & Ours & \textbf{0.25} & \textbf{0.97} & \textbf{0.98} & \textbf{1.00} & \textbf{1.00} & \textbf{0.77} & 4.43\\ \midrule
     \multirow{2}{*}{\begin{tabular}{c} Dormitory \end{tabular}} 
	 & SP+LG(Depth) & 0.05 & 0.33 & 0.49 & 2.63 & 6.94 & 2.22 & 4.20\\
     & Ours & \textbf{0.54} & \textbf{0.79} & \textbf{0.85} & \textbf{1.18} & \textbf{1.39} & \textbf{0.98} & \textbf{3.61}\\ \bottomrule
    \end{tabular}
        \begin{tablenotes}
            \footnotesize
            \item[1] SP+LG represents using SuperPoint\cite{detone2018superpoint} and LightGlue\cite{lindenberger2023lightglue} once in fine relocalization while the coarse retrieval stays the same with the proposed method. Depth means the depth-channel map image.
            \item[2] The RMSE, MSE, MAE, and Max Error are calculated based on correct relocalized queries, included in the category of RR(3m/5\textdegree).
        \end{tablenotes}
    \end{threeparttable}
    
    \label{fine}
\end{table*}

\section{Conclusion}
In this paper, we proposed a hierarchical visual cross-modal relocalization system utilizing texture consistency between the grayscale and intensity values. The proposed method was evaluated on self-collected datasets, showing the effectiveness of the proposed method with high recall of both place recognition and pose estimation. Taking advantage of intensity as texture, the proposed method is expected to be a promising solution for cross-modal visual relocalization in prior LiDAR maps. 


In future work, we are eager to further explore the intensity's role in cross-modal relocalization. For example, re-train the global retrieval and local matching network to learn the consistency more tightly. To alleviate the influence of sparsity, endow the network with the ability to utilize both geometry and texture information.

\bibliography{reference.bib}

\end{document}